\documentclass{article}
\usepackage{ijcai23}

\usepackage{times}
\usepackage{soul}
\usepackage{url}
\usepackage[hidelinks]{hyperref}
\usepackage[utf8]{inputenc}
\usepackage[small]{caption}
\usepackage{graphicx}
\usepackage{amsmath}
\usepackage{amsthm}
\usepackage{booktabs}
\usepackage{algorithm}
\usepackage{algorithmic}
\usepackage[switch]{lineno}

\usepackage{enumitem}    
\usepackage{multirow}
\usepackage{float}
\usepackage{amsmath}
\usepackage{comment}
\usepackage{bbding}
\usepackage{booktabs}    

\title{Adversarial Amendment is the Only Force\\Capable of
Transforming an Enemy into a Friend}

\author{
Chong Yu$^1$
\and
Tao Chen$^{2,*}$\And
Zhongxue Gan$^{1,}$\thanks{Tao Chen and Zhongxue Gan are corresponding authors.}
\affiliations
$^1$Academy for Engineering and Technology, Fudan University\\
$^2$School for Information Science and Technology, Fudan University
\emails
21110860050@m.fudan.edu.cn,
\{eetchen, ganzhongxue\}@fudan.edu.cn
}

\begin{document}

\maketitle

\begin{abstract}
    Adversarial attack is commonly regarded as a huge threat to neural networks because of misleading behavior. This paper presents an opposite perspective: adversarial attacks can be harnessed to improve neural models if amended correctly. Unlike traditional adversarial defense or adversarial training schemes that aim to improve the adversarial robustness, the proposed adversarial amendment (\textit{AdvAmd}) method aims to improve the original accuracy level of neural models on benign samples. We thoroughly analyze the distribution mismatch between the benign and adversarial samples. This distribution mismatch and the mutual learning mechanism with the same learning ratio applied in prior art defense strategies is the main cause leading the accuracy degradation for benign samples. The proposed \textit{AdvAmd} is demonstrated to steadily heal the accuracy degradation and even leads to a certain accuracy boost of common neural models on benign classification, object detection, and segmentation tasks. The efficacy of the \textit{AdvAmd} is contributed by three key components: \textbf{mediate samples} (to reduce the influence of distribution mismatch with a fine-grained amendment), \textbf{auxiliary batch norm} (to solve the mutual learning mechanism and the smoother judgment surface), and \textbf{\textit{AdvAmd} loss} (to adjust the learning ratios according to different attack vulnerabilities) through quantitative and ablation experiments.
\end{abstract}

\section{Introduction}
\label{sec:introduction}
The success of neural networks has brought a radical improvement in applications for human beings' daily life. Meanwhile, the concerns about the robustness and safety of the neural networks also increase.~\cite{szegedy2013intriguing} proved that by maximizing the 
prediction error, the neural model will always classify an image to a misleading category by applying a certain hardly perceptible perturbation. What is worse, this is not a random artifact of the neural network learning process. In contrast, this is a common issue
. And the perturbed inputs are termed ``adversarial samples". Since then, much attention has been put on the study of the adversarial attack, which is a technique that attempts to fool neural networks with deceptive data~\cite{goodfellow2014explaining}.

\textbf{\textit{Adversarial attack}} is commonly regarded as a growing and massive threat to the industry and research community because of its misleading behavior toward neural networks. With the in-depth study, adversarial attacks are proved to be effective in flexibly generating adversarial examples to deceive a range of applications, especially in the computer vision field, such as classification~\cite{moosavi2016deepfool}, semantic segmentation~\cite{xu2020adversarial}, face recognition~\cite{mirjalili2017soft}, and depth estimation~\cite{zhang2020adversarial}, also in the natural language processing field for 
recognition~\cite{cisse2017houdini}, 
generation~\cite{liang2017deep}, and 
translation~\cite{belinkov2017synthetic} tasks.
Various \textbf{\textit{adversarial defense}} strategies have been proposed, including modifying input data~\cite{tramer2017ensemble}~\cite{dziugaite2016study}, modifying to regularize neural models~\cite{papernot2016distillation}~\cite{lyu2015unified}, and using auxiliary tools~\cite{meng2017magnet}~\cite{samangouei2018defense} to improve the robustness and safety of 
models against various adversarial attacks. \textbf{Our method belongs to neither area}, i.e., \textit{it first applies the adversarial attack to generate attack samples, then amends the 
attack samples and the adversarial defense process to achieve robustness and boost the accuracy on the original benign dataset, simultaneously.}

The typical purpose of the adversarial attack is to add a natural perturbation to the benign input samples to generate the corresponding adversarial samples, which may cause a specific malfunction in the target model. Meanwhile, the adversarial perturbation is hardly perceptible, so humans can still correctly recognize the adversarial samples. 
The principle for most prior art adversarial attacks is changing the gradient step to a misleading direction when calculating the loss function during the back-propagation. 
The naive defense strategy is generating such adversarial samples and mixing them with benign samples to perform the mutual learning for neural models and make them more robust to adversarial attacks.
However, such mutual learning is based on the \textbf{\textit{assumption}} that \textbf{\textit{the adversarial and benign samples have a similar distribution}}, which is not necessarily correct, as observed from the 
experiment results in Table~\ref{Table.1}
. Because if the assumption stands, both the model accuracy metric on adversarial and metric on benign samples should increase. But in fact, we find such a defense strategy has a side-effect on the model capacity on benign samples, i.e., \textbf{\textit{leading to noticeable accuracy degradation on the benign dataset}}.

Inspired by the findings that the defense against adversarial attack (details in Section 3) may lead to the accuracy degradation on benign samples, this paper focuses on solving this issue. We stand on an opposite perspective to previous studies, i.e., if the adversarial attack can be amended properly and in the right direction, \textbf{\textit{the attack can be harnessed and transferred to improve the neural models' accuracy}}. Besides the similarity in the workflow, the \textbf{\textit{purpose}} of our technique is totally different from traditional adversarial defense or adversarial training methods, which aim at improving the adversarial robustness, i.e., the accuracy after introducing adversarial samples. In contrast, our proposed adversarial amendment method aims to improve the original capabilities of neural models, i.e., the accuracy induced by benign samples. Our main contributions include:
\begin{itemize}
    \item The qualitative explanation and theoretical proof illustrate the distribution mismatch 
    and the mutual learning mechanism cause the accuracy degradation 
    for benign samples. (Section 3) 
    \item 
    \textit{AdvAmd} 
    is featured with involving mediate samples, inserting auxiliary batch norm, and applying \textit{AdvAmd} loss. It solves the mismatch and tangled distribution with fine-grained data argumentation and learning adjustment according to attack vulnerabilities. (Section 4
    )
    \item We validate the efficacy of \textit{AdvAmd} both on benign and adversarial samples. On several classification, detection and segmentation models, and results show that our  \textit{AdvAmd} method can achieve about $\textbf{1.2\%}\sim\textbf{2.5\%}$ metric increase to on benign samples without side-effect on robustness against adversarial samples
    . (Section 5
    )
\end{itemize}


\section{Related Work}
\label{sec:related_work}
In our proposed method, we need to apply the adversarial attack to generate the adversarial samples, so we first go through the standard adversarial attack methods. Though adversarial defense has a different purpose from the proposed amendment, it has a similar workflow. So we also review the adversarial defense methods for further distinction.

\subsection{Adversarial Attack}
\label{subsec:adversarial_attack}
Based on whether or not having access to the target model, the adversarial attack methods~\cite{ozdag2018adversarial} can be divided into the black-box and the white-box categories. Because our aim is amending the adversarial attack to boost the accuracy of the original model, the assumption is having access to the model which belongs to the white-box attack category.

\textbf{White-box attack.} In 
this category, the full knowledge of the target model, including architecture, parameters, training method and dataset, is assumed to be known. And the adversarial attacker can fully utilize the available information to analyze the most vulnerable points of the target model.
~\cite{goodfellow2014explaining} proposed the Fast Gradient Sign Method (\textit{FGSM}), which calculates the gradient of the cost function during the back-propagation, then generates the adversarial examples by changing one gradient sign step. 
Basic Iterative Method (\textit{BIM})~\cite{kurakin2016adversarial} and Projected Gradient Descent (\textit{PGD})
~\cite{madry2018towards} are the straightforward extension of \textit{FGSM} by applying a \textit{FGSM} attack multiple times with a small step size.
~\cite{moosavi2016deepfool} proposed a \textit{DeepFool} attack to compute a minimal norm 
perturbation 
in an iterative manner to find the decision boundary and find the minimal adversarial samples across the boundary.~\cite{cisse2017houdini} proposed a \textit{Houdini} attack, which is demonstrated to be effective in generating perturbations to deceive gradient-based networks used in image classification and speech recognition tasks.
In our amendment workflow, we mainly apply \textit{FGSM}, \textit{PGD} and \textit{DeepFool} to generate the adversarial samples.

\subsection{Adversarial Defense}
\label{subsec:adversarial_defense}
A mass of adversarial defense methods have been proposed to improve the robustness of neural models against the adversarial attack, which can be divided into three main categories: modifying data, modifying models, and using auxiliary tools. The defense strategy in the first category does not directly deal with the target models. In contrast, the other two categories are more concerned with the target models themselves.

\textbf{Data modification.} Adversarial training is the most widely used method in this category.~\cite{szegedy2013intriguing} injected adversarial samples and modified their labels to improve the robustness of the target model.~\cite{huang2015learning} increased the robustness of the target model by punishing misclassified adversarial samples. The limitation of this strategy is that if all unknown adversarial samples are introduced into the training 
then the accuracy will be decreased 
on benign samples. In contrast, introducing some of the adversarial samples is often not enough to remove the impact of the adversarial perturbation. The \textbf{\textit{mediate samples}} proposed in our amendment method can provide fine-grained data modification as a fix.

\textbf{Model modification.} 
The popular strategy is defensive distillation. This kind of strategy extends the knowledge distillation~\cite{hinton2015distilling} to producing a new target model with a smoother output surface~\cite{papernot2016distillation}~\cite{papernot2017extending}, that is less sensitive to adversarial perturbations to improve the robustness. However, the smoother output surface may also lead the model to make ``new" mistakes when detecting the benign samples, which is proved in Section 3
. It inspires us to apply \textbf{\textit{two separate batch norms}} for learning from benign and adversarial samples, which can improve robustness and simultaneously keep the sharper output surface.

\textbf{Auxiliary tools utilization.} Researchers also came up with defense strategies by using auxiliary tools.
The successful strategy is proposed as \textit{MagNet}~\cite{meng2017magnet}. \textit{MagNet} uses an auxiliary detector to identify the benign and adversarial samples by measuring the distance between a given test sample and the manifold, then rejects the sample if the distance exceeds the threshold. However, based on our experiments in Table~\ref{Table.1}, it still leads the accuracy degradation on benign samples. It inspires us the necessity to \textbf{\textit{adjust the learning ratios}} for different samples \textbf{\textit{according to different attack vulnerabilities}}.


\section{Existing Defense Strategies Lead to Accuracy Degradation On Benign Samples}
\label{sec:fail_to_boost_accuracy}
The existing adversarial defense strategies are effective and helpful to target models' adversarial robustness, i.e., boosting the target models' accuracy when testing on adversarial samples. 
Some prior works find these strategies may lead to the accuracy degradation on benign samples~\cite{meng2017magnet}~\cite{madry2018towards}~\cite{xie2019feature} for the basic classification task. Because object detection is 
more complicated than classification, we want to verify whether such accuracy degradation is also valid in the object detection task. We validate by choosing the representative defense method in each category (adversarial training~\cite{szegedy2013intriguing} (\textit{Adv-Train}), defensive distillation~\cite{papernot2017extending} (\textit{Def-Distill}), and \textit{MagNet}~\cite{meng2017magnet}) and apply them to several typical object detection models, i.e., \textit{Faster R-CNN}~\cite{ren2015faster}, \textit{SSD}~\cite{liu2016ssd}, \textit{RetinaNet}~\cite{lin2017focal}, \textit{YOLO}~\cite{glenn_jocher_2022_6222936}) to test the corresponding accuracy on the benign and adversarial \textit{COCO} dataset~\cite{lin2014microsoft}. The adversarial dataset is generated by \textit{FGSM}~\cite{goodfellow2014explaining} and \textit{PGD}~\cite{madry2018towards} attacks. 
More details can refer to \textbf{Experiments Settings}.

\begin{table}[htb]
\centering
\resizebox{0.995\linewidth}{!}{
\begin{tabular}{lllcccccccc}
\toprule
\multirow{3}{*}{\textbf{Network}} & \multirow{3}{*}{\textbf{\begin{tabular}[c]{@{}c@{}}Baseline\\ Box AP\end{tabular}}} & \multirow{3}{*}{\textbf{\begin{tabular}[c]{@{}c@{}}Defense\\ Type\end{tabular}}} & \multicolumn{4}{c}{\textbf{$\Delta$ Box AP on Adversarial Dataset}} & \multicolumn{4}{c}{\textbf{$\Delta$ Box AP on Benign Dataset}} \\
\cmidrule(lr){4-7}\cmidrule(lr){8-11}
 & & & \multicolumn{2}{c}{\textbf{FGSM Attack}} & \multicolumn{2}{c}{\textbf{PGD Attack}} & \multicolumn{2}{c}{\textbf{FGSM Attack}} & \multicolumn{2}{c}{\textbf{PGD Attack}}\\
\cmidrule(lr){4-5}\cmidrule(lr){6-7}\cmidrule(lr){8-9}\cmidrule(lr){10-11}
 & & & \multicolumn{1}{c}{\textbf{$\epsilon$=0.01}} & \multicolumn{1}{c}{\textbf{$\epsilon$=0.1}} & \multicolumn{1}{c}{\textbf{$\epsilon$=0.01}} & \multicolumn{1}{c}{\textbf{$\epsilon$=0.1}} & \multicolumn{1}{c}{\textbf{$\epsilon$=0.01}} & \multicolumn{1}{c}{\textbf{$\epsilon$=0.1}} & \multicolumn{1}{c}{\textbf{$\epsilon$=0.01}} & \multicolumn{1}{c}{\textbf{$\epsilon$=0.1}} \\
\midrule
\multirow{4}{*}{\begin{tabular}[c]{@{}l@{}}Faster R-CNN\\ (RN50)\end{tabular}}    & \multirow{4}{*}{\textbf{37.0}} 
   & None        & -5.1  & -9.3  & -6.5  & -11.1  & 0.0   & 0.0   & 0.0   & 0.0  \\
 & & Adv-Train   & -2.8  & -5.4  & -3.8  & -6.8   & -1.3  & -3.0  & -2.0  & -3.2 \\
 & & Def-Distill & -2.6  & -5.3  & -3.8  & -6.7   & -1.4  & -3.0  & -2.0  & -3.3 \\
 & & MagNet      & -2.2  & -5.4  & -3.7  & -6.6   & -1.7  & -3.2  & -2.3  & -3.5 \\
\midrule
\multirow{4}{*}{\begin{tabular}[c]{@{}l@{}}SSD\\ (RN50)\end{tabular}}    & \multirow{4}{*}{\textbf{25.8}} 
   & None        & -4.7  & -8.7  & -5.6  & -9.8  & 0.0   & 0.0   & 0.0   & 0.0  \\
 & & Adv-Train   & -2.4  & -5.4  & -3.9  & -6.2  & -0.7  & -1.8  & -1.4  & -2.1 \\
 & & Def-Distill & -2.0  & -5.1  & -3.6  & -6.0  & -0.8  & -1.9  & -1.5  & -2.3 \\
 & & MagNet      & -1.8  & -4.9  & -3.5  & -5.9  & -0.9  & -2.1  & -1.5  & -2.4 \\
\midrule
\multirow{4}{*}{\begin{tabular}[c]{@{}l@{}}RetinaNet\\ (RN50)\end{tabular}}    & \multirow{4}{*}{\textbf{36.4}} 
   & None        & -5.0  & -9.1  & -6.4  & -10.9  & 0.0   & 0.0   & 0.0   & 0.0  \\
 & & Adv-Train   & -2.7  & -5.3  & -3.7  & -6.7   & -1.3  & -2.9  & -1.9  & -3.1 \\
 & & Def-Distill & -2.5  & -5.2  & -3.7  & -6.6   & -1.4  & -3.0  & -1.9  & -3.2 \\
 & & MagNet      & -2.2  & -5.3  & -3.6  & -6.5   & -1.6  & -2.9  & -2.0  & -3.4 \\
\midrule
\multirow{4}{*}{\begin{tabular}[c]{@{}l@{}}YOLO-V5\\ (Large)\end{tabular}}    & \multirow{4}{*}{\textbf{48.6}} 
   & None        & -6.9  & -12.2  & -8.5  & -14.8  & 0.0   & 0.0   & 0.0   & 0.0  \\
 & & Adv-Train   & -3.8  & -6.4   & -4.7  & -7.9   & -2.2  & -4.5  & -3.2  & -5.3 \\
 & & Def-Distill & -3.6  & -6.3   & -4.5  & -7.6   & -2.4  & -4.6  & -3.4  & -5.4 \\
 & & MagNet      & -3.4  & -6.2   & -4.4  & -7.6   & -2.5  & -4.8  & -3.5  & -5.5 \\
\bottomrule
\end{tabular}
}
\caption{Adversarial defense strategies effectiveness on detection models. ($\epsilon$: perturbation epsilon)}
\label{Table.1}
\vskip -0.05in
\end{table}

From the results shown in Table~\ref{Table.1}, we can draw two conclusions. \textbf{\textit{The adversarial defense methods effectively improve the adversarial robustness}}, i.e., the accuracy of the adversarial dataset has been obviously recovered. On the other hand, \textbf{\textit{the accuracy degradation on the benign dataset is also valid for the detection models}}. The detection models ``enhanced" with defense methods obtain lower accuracy on benign datasets than their vanilla baselines. If the adversarial dataset is attacked with more strength and generated with more perturbations, the accuracy degradation will be more noticeable when passing the adversarial defense ``enhancement". Therefore, the diverse behaviors on the adversarial and benign dataset prove the assumption that the adversarial and benign samples have a similar distribution does not stand. \textit{So the mutual learning on benign and adversarial samples in the adversarial defense methods cannot avoid the accuracy degradation on the benign samples if they also want to keep the high adversarial robustness.} This dilemma is the key issue to be solved in this paper.

\subsection{Qualitative Explanation}
\label{subsec:qualitative_explanation}
We hypothesize such accuracy degradation for benign samples is mainly caused by distribution mismatch and fuzz. Though the adversarial perturbations are slight in magnitude, the distribution of the adversarial samples differs from the benign counterparts. The essential concepts of adversarial training and \textit{MagNet} are both harnessing the adversarial samples with the corrected label information in the corresponding regions in the benign samples, then involving these processed samples with vanilla benign samples in adversarial training. The supervised training samples from two different sets with different distributions will make the judgment boundary fuzzy. The defensive distillation aims to produce an enhanced target model with a smoother output surface, also eventually fuzzing the boundary. With the vague boundary, the target model will make fewer mistakes when detecting the adversarial samples but may also make ``new" mistakes when detecting the benign samples.

\begin{figure*}[htb]
\vskip -0.10in
\begin{center}
    \centering
    \begin{minipage}[b]{0.325\linewidth}
		\centering
		\textbf{\texttt{\fontsize{7.5pt}{\baselineskip}\selectfont (a) Vanilla Object Detection}}
	\end{minipage}
	\begin{minipage}[b]{0.325\linewidth}
		\centering
		\textbf{\texttt{\fontsize{7.5pt}{\baselineskip}\selectfont (b) Adversarial Attack}}
	\end{minipage}
	\begin{minipage}[b]{0.325\linewidth}
		\centering
		\textbf{\texttt{\fontsize{7.5pt}{\baselineskip}\selectfont (c) Adversarial Defense}}
	\end{minipage}
\vskip 0.04in
    \begin{minipage}[b]{0.325\linewidth}
		\centering
		\includegraphics[width = 0.99\linewidth]{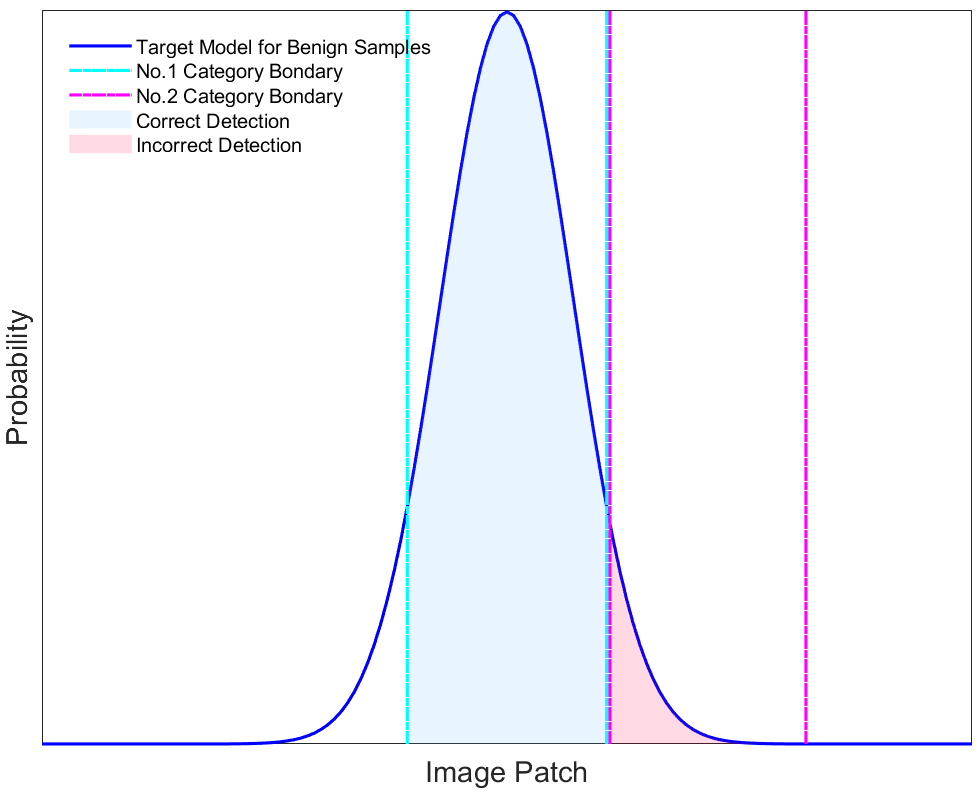}
	\end{minipage}
	\begin{minipage}[b]{0.325\linewidth}
		\centering
		\includegraphics[width = 0.99\linewidth]{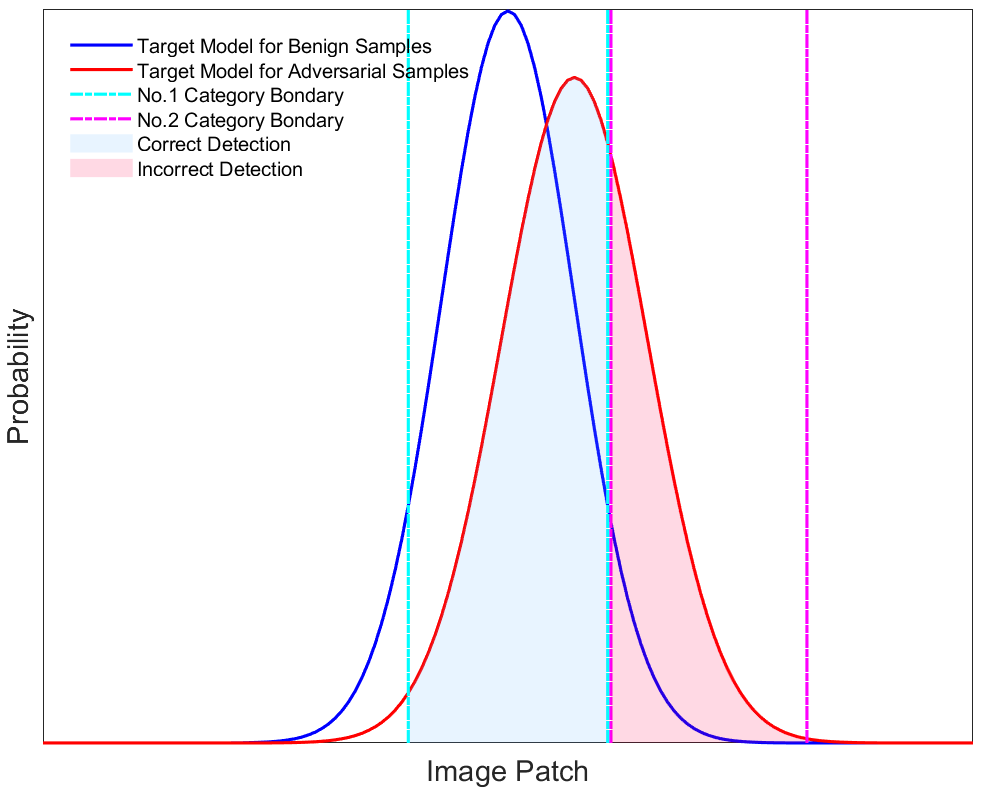}
	\end{minipage}
	\begin{minipage}[b]{0.325\linewidth}
		\centering
		\includegraphics[width = 0.99\linewidth]{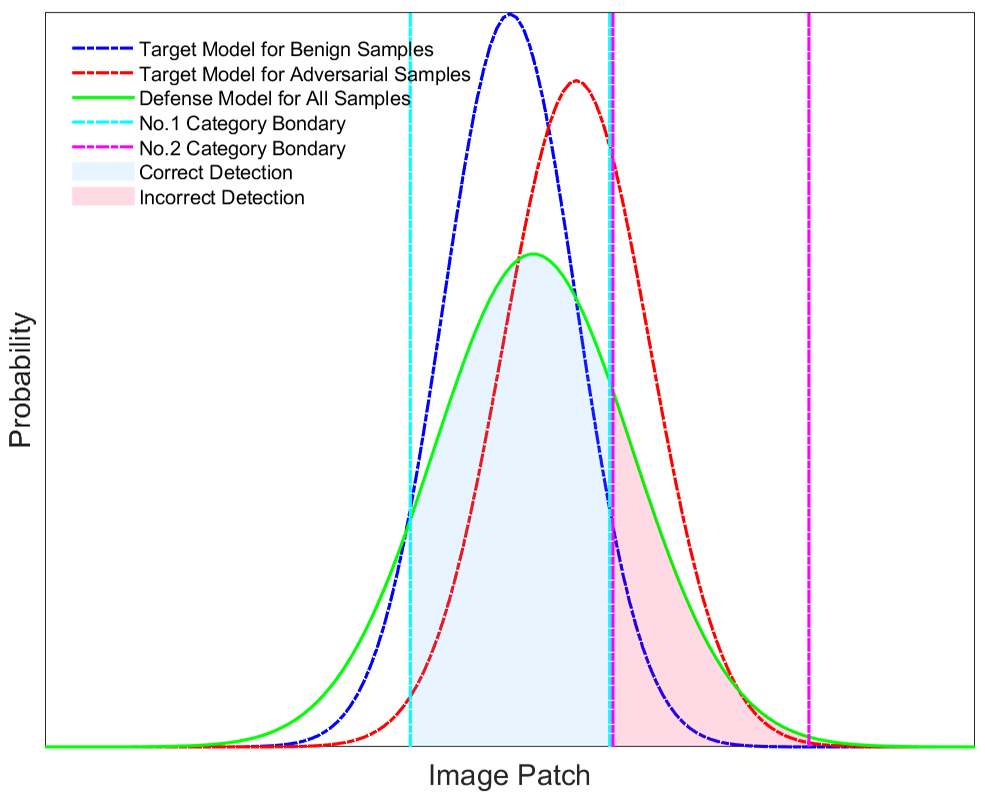}	
	\end{minipage}
\end{center}
\vskip -0.15in
	\caption{Qualitative explanation for adversarial attack and defense influence
 . Comparison between (a) and (b) helps to explain why the model is misleading by adversarial samples. Comparison between (b) and (c) helps to explain why the defense model has better adversarial robustness to the adversarial samples. Comparison between (a) and (c) helps to explain the accuracy degradation for benign samples.}
	\label{Figure.1}
\end{figure*}

We illustrate the qualitative explanation based on the hypothesis, as shown in Figure~\ref{Figure.1}. To be clear, we only pay close attention to two object categories divided and marked by the boundaries in the figure. For a vanilla-trained object detection model, its judgment distribution to the benign samples is shown in Figure~\ref{Figure.1} (a). The corresponding area filled with light blue background color represents the model's correct detection for the No.1 category. In contrast, the area filled with light red represents the incorrect detection, i.e., detecting the objects belonging to the No.1 category as the No.2 category. As shown in Figure~\ref{Figure.1} (b), \textit{when the adversarial attack perturbs the benign samples, the distribution of the adversarial samples drifts from the benign counterparts, and so do the correct and incorrect detection areas.} The incorrect detection area size is obviously enlarged, which aligns with the phenomenon that the model is misleading by some adversarial samples. Then the adversarial defense method is applied, as shown in Figure~\ref{Figure.1} (c). Based on the explanation mentioned above, the label info correction in each defense strategy will not alter the model judgment distribution before the defense process. The defense training process obtains info from both benign and adversarial distribution, producing a defense model with a smoother judgment distribution. Compared to Figure~\ref{Figure.1} (b), the incorrect detection area is diminished, which aligns with the phenomenon that the defense model has better adversarial robustness to the adversarial samples. We also notice the right detection area size is smaller than the counterpart in Figure~\ref{Figure.1} (a) due to the smoother judgment surface. That helps explain why the accuracy degradation for benign samples occurs qualitatively.


\subsection{Theoretical Proof}
\label{subsec:theoretical_proof}

Starting from the two-category detection situation, assume the aforementioned vanilla-trained model's judgment distributions to the benign and adversarial samples are two normal random variables $X$, $Y$ with means $\mu_{x}$, $\mu_{y}$ and variances $\sigma_{x}^{2}$, $\sigma_{y}^{2}$., which can be expressed as follows:
\begin{equation}
X \sim N\left (\mu_{x}, \sigma_{x}^{2} \right ), \quad Y \sim N\left (\mu_{y}, \sigma_{y}^{2} \right ) 
\end{equation}

\textbf{Theorem:}\label{theorem2} Mutual probability distribution of the linear combination of multiple normal random variables: $X_1 \sim N\left (a_1\mu_{1}, A_1^2{\sigma_{1}}^{2} \right )$, $X_2 \sim N\left (a_2\mu_{2}, A_2^2{\sigma_{2}}^{2} \right )$ $\cdots X \sim N\left (a_n\mu_{n}, A_n^2{\sigma_{n}}^{2} \right )$ can be expressed as: $Z=\sum_{i=1}^{n}c_iX_i \sim N\left (\sum_{i=1}^{n}c_ia_i\mu_{i}, \sum_{i=1}^{n}c_i^2A_i^2{\sigma_{i}}^{2} \right )$.

From the 
\textbf{Theorem} we can find the mutual probability distribution has larger variances than each variance for each individual distribution\footnote{If the multiple normal
random variables has close correlations, the mutual probability distribution is more complicated. Expressions can refer to proof~\cite{nadarajah2016distribution}. However the conclusion for larger variance will not change.}. And the larger variance value leads to the smoother judgement boundary for the defense model, which helps explain why the accuracy degradation for benign samples occurs in theory.


\section{Adversarial Amendment is Capable of Transforming an Enemy into a Friend}
\label{sec:adversarial_amendment}
Based on the qualitative explanation and theoretical proof, we find the distribution mismatch between the benign and adversarial samples, and the mutual learning mechanism widely applied in various adversarial defense strategies cause the accuracy degradation for benign samples. In this section, we propose the \textbf{Adv}ersarial \textbf{Am}en\textbf{d}ment (\textit{AdvAmd}) method to harness the adversarial attack and samples properly,  \textbf{\textit{transforming these commonly regarded enemies into a friend}}, i.e., \textit{improving 
model accuracy on benign samples}.

Given the supervised category info $\textbf{\emph{y}}_c$, the vanilla training and optimization process of 
target 
model (referred as \begin{small}$\textbf{\emph{M}}_T$\end{small}) can be expressed as the following minimization problem:
\begin{equation}
\min_{\theta}\ \textbf{\emph{L}}(\textbf{\emph{y}}, \textbf{\emph{y}}_c)=
\min_{\theta}\ \textbf{\emph{L}}(\textbf{\emph{F}}(\textbf{\emph{M}}_T, \textbf{\emph{x}}), \textbf{\emph{y}}_c)
\end{equation}
where $\textbf{\emph{x}}$ is the benign input sample, $\textbf{\emph{y}}$ is the corresponding output of the neural function \begin{small}$\textbf{\emph{F}}$\end{small}, $\theta$ is the weight parameters of the target model \begin{small}$\textbf{\emph{M}}_T$\end{small}, and \begin{small}$\textbf{\emph{L}}(\cdot)$\end{small} refers to the loss function used in vanilla training.

For the adversarial attack settings, the perturbation $\delta$ is added to the input $\textbf{\emph{x}}$ to constitute the adversarial samples $\textbf{\emph{x}}_{adv}$, 
subject to the $L_{*}$ perturbation constraint:
\begin{equation}
\left \| \textbf{\emph{x}}_{adv} - \textbf{\emph{x}} \right \|_{*} = \left \| \delta \right \|_{*} \leq \epsilon, \quad *\in \left \{ 0,1,2,\infty \right \}
\end{equation}
where $\epsilon$ refers to the attack strength. The aim for the adversarial attack is to maximally deteriorate the benign samples and change the correct 
behavior of the target model within the perturbation constraint. If the whole benign dataset is defined as \begin{small}$\textbf{\emph{D}}$\end{small}, with \begin{small}$N$\end{small} categories, and each category is labeled with $\textbf{\emph{y}}_{c:k}$ and the subset with category $\textbf{\emph{y}}_{c:k}$ is denoted as \begin{small}$\textbf{\emph{D}}(\textbf{\emph{y}}_{c:k})$\end{small}, then the adversarial attack process can be denoted with the following joint optimization problem.
\begin{equation}
\label{adversarial_attack_joint}
\left\{\begin{matrix}
\begin{aligned}
&\mathop{\max_{\delta_k}}\sum\limits_{k = 1}^{N}\sum\limits_{\textbf{\emph{x}}_k \in \textbf{\emph{D}}(\textbf{\emph{y}}_{c:k})}\textbf{\emph{L}}(\textbf{\emph{F}}(\textbf{\emph{M}}_T, \textbf{\emph{x}}_k + \delta_k), \textbf{\emph{y}}_{c:k}) \\
&\mathop{\min}\sum\limits_{k = 1}^{N}\left \| \delta_k \right \|_*, \quad \left \| \delta_k \right \|_* \leq \epsilon, \quad *\in \left \{ 0,1,2,\infty \right \} \\
\end{aligned}
\end{matrix}\right.
\end{equation}

\noindent \textbf{Pain point 1:}\label{pain_point1} distribution mismatch between the benign and adversarial samples.

In the existing adversarial defense strategies, the supervised label 
of the adversarial sample is corrected by the corresponding region in the benign samples. However, the label 
correction in each defense strategy will not alter the distribution for these adversarial samples. So the distribution mismatch still exists in the following defense process. \textbf{\textit{The \textit{AdvAmd} method controls the prepositive adversarial attack in fine-granularity.}} The adversarial perturbations are added and alter the distribution of the samples in iterations. In general, the more attack iterations and the larger perturbations lead to more distribution drift. The \textit{AdvAmd} method 
collects the mediate samples in the attack process to generate the successful adversarial samples. These mediate samples are inside the boundary to mislead the target model so that we can use the label info in the corresponding benign samples. Based on the expression (\ref{adversarial_attack_joint}) for adversarial attack, mediate samples referred as $\textbf{\emph{x}}_{med}$ can be expressed as:
\begin{equation}
\label{pain_point1_equation}
\left \{\textbf{\emph{x}}_{med} = \textbf{\emph{x}}_k + \varphi \delta_k|\textbf{\emph{x}}_k \in \textbf{\emph{D}}(\textbf{\emph{y}}_{c:k}), k=1, \cdots N \right \}
\end{equation}
where $\varphi \in \left ( 0,1\right )$ refers to the mediate coefficient.

\noindent \textbf{Pain point 2:}\label{pain_point2} mutual learning mechanism applied in various adversarial defense strategies leading to a smoother surface.

This solution to the first pain point can reduce the influence of distribution mismatch but cannot eliminate the differences between the adversarial and benign samples. The mutual learning from the mixture of 
adversarial and benign samples still generate a smoother judgment surface. 
Batch normalization~\cite{ioffe2015batch} (\textit{BN}) is an essential component for 
neural models. Specifically, \textit{BN} normalizes input samples by the mean and variance dynamically computed within each mini-batch. The \textit{BN} layers are especially effective when the input samples in each mini-batch have the same or similar distributions. These \textit{BN} layers lose efficacy when the input samples in one mini-batch come from totally different distributions, resulting in inaccurate statistics estimation and normalization. Inspired by the efficacy of \textit{BN}, \textbf{\textit{
\textit{AdvAmd} method further disentangles the mutual learning mechanism in the mixture distribution into two separate paths for the adversarial and benign samples, respectively.}} 
\textit{AdvAmd} method inserts an auxiliary \textit{BN} aside from each original \textit{BN} layer to guarantee 
normalization statistics are exclusively performed on the adversarial examples.

\noindent \textbf{Loss Enhancement:} The detection difficulties vary among the multiple object categories for the object detection task. Intuitively, if the object in a specific category is easy to be incorrectly detected as another category, it reflects this object is hard to detect. In contrast, if the object in a certain category is hard to be attacked by the adversarial perturbation, it means this object category is relatively easy to detect. So does the classification task. For multi-category 
task with \begin{small}$N$\end{small} categories, the \textit{AdvAmd} loss is defined as follows:
\setlength\abovedisplayskip{1pt}
\setlength\belowdisplayskip{1pt}
\begin{equation}
\label{AdvAmd_loss}
\begin{aligned}
&Loss_{AdvAmd}=Loss_{A}= -\sum\limits_{k = 1}^{N}\bar{\emph{A}}y_{o,k}log\left (p_{o,k} \right ), \\
&\bar{\emph{A}}= 1-\frac{1}{2(N-1)} \left (\sum\limits_{i \neq k,i = 1}^{N}\alpha_{k}^{i}+\sum\limits_{j \neq k,j = 1}^{N}\alpha_{j}^{k} \right )
\end{aligned}
\end{equation}
where $y_{o,k}$ refers to the binary indicator whether the category label $k$ is the correct detection result for observation $o$, $p_{o,k}\in \left [ 0,1\right ]$ is the model's estimated probability if the observation $o$ is detected as the category $k$. $\alpha_{k}^{i}\in \left [ 0,1\right ]$ refers to the attack difficulty from changing the model detection category for $k$ to $i$, and larger value means the higher attack difficulty. So the item \begin{small}$\sum_{i \neq k,i = 1}^{N}\alpha_{k}^{i}$\end{small} is the sum of the attack difficulties by changing the model detection category from a given category $k$ to all the other categories in the dataset, while the item \begin{small}$ \sum_{j \neq k,j = 1}^{N}\alpha_{j}^{k}$\end{small} is the sum of the attack difficulties by changing the model detection category from the other categories to a given category $k$. $\bar{\emph{A}}\in \left [ 0,1\right ]$ refers to the normalized adversarial attack vulnerable coefficient, and larger value of $\bar{\emph{A}}$ means the model is more vulnerable to the attack, i.e. lower attack difficulty.

\begin{algorithm}[!htb]
\algsetup{linenosize=\tiny}
\small
  \caption{\textbf{Adv}ersarial \textbf{Am}en\textbf{d}ment (\textit{AdvAmd})
  }
  \label{Algorithm}
  \textbf{Input}: Target 
  model $\textbf{\emph{M}}_T$ with neural function $\textbf{\emph{F}}$, Benign dataset $\textbf{\emph{D}}$ with $N$ categories. Each category is labeled with $\textbf{\emph{y}}_{c:k}$, and the subset with category $\textbf{\emph{y}}_{c:k}$ is denoted as $\textbf{\emph{D}}(\textbf{\emph{y}}_{c:k})$. \\
  \textbf{Parameter}: Attack strength $\epsilon$, Perturbation constraint $\left \| \cdot \right \|_* \left ( *\in \left \{ 0,1,2,\infty \right \} \right )$, Mediate coefficient $\varphi \in \left ( 0,1\right )$, Loss adjustment factors $\beta_1, \beta_2, \beta_3$, Overall loss threshold $\sigma$. \\
  \textbf{Output}: Amended model $\textbf{\emph{M}}_A$. \\
  \begin{algorithmic}[1] 
    \FOR{benign samples $\textbf{\emph{x}}$ in the benign dataset $\textbf{\emph{D}}$}
    \STATE \textbf{\emph{Adversarial Attack Process}} by optimizing the adversarial perturbation $\delta_k$:
    \STATE 
    $
    \qquad\left\{\begin{matrix}
    \begin{aligned}
    &\mathop{\max\limits_{\delta_k}}\sum\limits_{k = 1}^{N}\sum\limits_{\textbf{\emph{x}}_k \in \textbf{\emph{D}}(\textbf{\emph{y}}_{c:k})}\textbf{\emph{L}}(\textbf{\emph{F}}(\textbf{\emph{M}}_T, \textbf{\emph{x}}_k + \delta_k), \textbf{\emph{y}}_{c:k}) \\
    &\mathop{\min}\sum\limits_{k = 1}^{N}\left \| \delta_k \right \|_*, \quad \left \| \delta_k \right \|_* \leq \epsilon \\
    \end{aligned}
    \end{matrix}\right.
    $
    \STATE Generate adversarial samples $\textbf{\emph{x}}_{adv}$: \\
    $\qquad \left \{\textbf{\emph{x}}_{adv} = \textbf{\emph{x}}_k + \delta_k|\textbf{\emph{x}}_k \in \textbf{\emph{D}}(\textbf{\emph{y}}_{c:k}), k=1, \cdots N \right \}$
    \STATE Generate mediate samples $\textbf{\emph{x}}_{med}$: \\
    $\qquad \left \{\textbf{\emph{x}}_{med} = \textbf{\emph{x}}_k + \varphi \delta_k|\textbf{\emph{x}}_k \in \textbf{\emph{D}}(\textbf{\emph{y}}_{c:k}), k=1, \cdots N \right \}$
    \STATE Generate adversarial attack vulnerable coefficient $\bar{\emph{A}}$: \\
    $\qquad \bar{\emph{A}} = 1-\frac{1}{2(N-1)} \left (\sum\limits_{i \neq k,i = 1}^{N}\alpha_{k}^{i}+\sum\limits_{j \neq k,j = 1}^{N}\alpha_{j}^{k} \right )
    $
    \ENDFOR
  \STATE Init Amended model $\textbf{\emph{M}}_A$ with the original target 
  model $\textbf{\emph{M}}_T$.
  \WHILE{overall loss: $Loss_{Overall} > \sigma$}
    \FOR{benign samples $\textbf{\emph{x}}$ in the benign dataset $\textbf{\emph{D}}$}
    \STATE Get the corresponding adversarial samples $\textbf{\emph{x}}_{adv}$ and mediate samples $\textbf{\emph{x}}_{med}. $
    \STATE Compute loss on benign samples using the original \textit{BN} layers as target model $\textbf{\emph{M}}_T$: \\
    $\qquad 
    Loss_{B}= -\sum\limits_{k = 1}^{N}\textbf{\emph{F}}(\textbf{\emph{M}}_A, \textbf{\emph{x}})_{o,k}log\left (p_{o,k} \right )$
    \STATE Compute loss on mediate samples using the original \textit{BN} layers as target model $\textbf{\emph{M}}_T$: \\
    $\qquad 
    Loss_{M}= -\sum\limits_{k = 1}^{N}\textbf{\emph{F}}(\textbf{\emph{M}}_A, \textbf{\emph{x}}_{med})_{o,k}log\left (p_{o,k} \right )$
    \STATE Compute loss on adversarial samples using the auxiliary \textit{BN} layers added in $\textbf{\emph{M}}_A$: \\
    $\qquad 
    Loss_{A} = -\sum\limits_{k = 1}^{N}\bar{\emph{A}}\textbf{\emph{F}}(\textbf{\emph{M}}_A, \textbf{\emph{x}}_{adv})_{o,k}log\left (p_{o,k} \right )$
    \STATE Calculate the overall loss: \\
    $\qquad Loss_{O} =\beta_1*Loss_{B} + \beta_2*Loss_{M} + \beta_3*Loss_{A}$
    \ENDFOR
    \STATE Minimize the overall loss w.r.t. parameters in amended model $\textbf{\emph{M}}_A$: $\min\ Loss_{Overall}$
  \ENDWHILE
  \STATE \textbf{return} Amended model $\textbf{\emph{M}}_A$ generated by \textit{AdvAmd} method.
  \end{algorithmic}
\end{algorithm}

Combining the improvements listed above, we formally propose the \textit{AdvAmd} workflow in Algorithm~\ref{Algorithm} to harness the adversarial attack and transfer to improve the object detection models' accuracy in the benign samples. In the first stage of the \textit{AdvAmd} method, a fine-grained adversarial attack is processed to generate the adversarial and mediate samples. Meanwhile, the adversarial attack vulnerable coefficients are calculated. Then we initialize the \textit{AdvAmd} amended model with the original network parameters of the target model. During the second stage of the \textit{AdvAmd} method, loss on the benign and mediate samples are calculated through the original \textit{BN} layers, while the adversarial samples need go through the auxiliary \textit{BN} layers. Finally, the weighted sum of three loss items is minimized with regard to the network parameter of the amended model for gradient updates.


\section{Experiments}
\label{sec:experiments}

\begin{table*}[!htb]
\resizebox{\linewidth}{!}{
\centering
\begin{tabular}{lcccccccc}
\toprule
Network  &  Optimizer  &  Initial LR  &  LR schedule  &  Momentum  &  Weight Decay  &  Epochs & Batch Size & GPU Num \\
\midrule
ResNet-50\textsuperscript{\ref{torchvision}}           & SGD            & 0.1   & Multi-Step (milestone: 30,60) & 0.9   & 1e-4 & 90 & 32 & 8 \\
DeiT-base\textsuperscript{\ref{deit}}                  & AdamW          & 0.0005 & Cosine Annealing              & 0.9   & 0.05 & 300 & 64 & 16 \\
\hline
Faster R-CNN (RN50)\textsuperscript{\ref{torchvision}} & SGD            & 0.02   & Multi-Step (milestone: 16,22) & 0.9   & 1e-4 & 26 & 4 & 8\\
SSD (RN50)\textsuperscript{\ref{nvidia_joc}}           & SGD            & 0.0026 & Multi-Step (milestone: 43,54) & 0.9   & 5e-4 & 65 & 32 & 8\\
RetinaNet (RN50)\textsuperscript{\ref{detectron2}}     & SGD            & 0.01   & Multi-Step (milestone: 16,22) & 0.9   & 1e-4 & 26 & 4 & 8 \\
EfficientDet-D0\textsuperscript{\ref{nvidia_joc}}      & Fused Momentum & 0.65   & Cosine Annealing              & 0.9   & 4e-5 & 300 & 60 & 8\\
YOLO-V5 (Large)\textsuperscript{\ref{yolov5}}          & SGD            & 0.01   & Linear                        & 0.937 & 5e-4 & 300 & 192 & 8 \\
\hline
Mask R-CNN (RN50)\textsuperscript{\ref{detectron2}}    & SGD            & 0.02   & Multi-Step (milestone: 16,22) & 0.9   & 1e-4 & 26 & 4 & 8\\
\bottomrule
\end{tabular}
} 
\caption{Experiments hyper-parameters for the classification, object detection and segmentation models tested in this paper.}
\label{Table.hyperparams}
\end{table*}

\subsection{Experiments Settings}
\label{subsec:experiments_settings}
For the experiments in this paper, we choose PyTorch~\cite{paszke2017automatic} with version 1.10.0 as the framework to implement all algorithms. All of the 
neural models, adversarial attack, defense and training, as well as \textit{AdvAmd} training and fine-tuning experimental results are obtained with V100~\cite{nvidiav100} and A100~\cite{nvidiaa100} GPU clusters. 
All the accuracy results reported for our proposed \textit{AdvAmd} method use FP32 
as the default data type. All the reference algorithms use the default data type provided in public repositories.

For the adversarial attack methods, we choose the \textit{FGSM}~\cite{goodfellow2014explaining}, \textit{DeepFool}~\cite{moosavi2016deepfool}, 
\textit{PGD}~\cite{madry2018towards} in Adversarial Robustness Toolbox\footnote{\url{https://github.com/Trusted-AI/adversarial-robustness-toolbox}\label{art}} as the reference algorithms. In the experiment, $L_{\infty}$ is chosen to measure the perturbation range and the attack strength.

To evaluate the effectiveness of the \textit{AdvAmd} and the other reference methods on the classification task, \textit{ResNet-50}\footnote{\url{https://github.com/pytorch/vision}\label{torchvision}}
~\cite{he2016deep} and \textit{DeiT}\footnote{\url{https://github.com/facebookresearch/deit}\label{deit}}~\cite{touvron2021training} are chosen as the experiment target models. For the detection task, \textit{Faster R-CNN}\textsuperscript{\ref{torchvision}}~\cite{ren2015faster}, \textit{SSD}\footnote{\url{https://github.com/NVIDIA/DeepLearningExamples}\label{nvidia_joc}}
~\cite{liu2016ssd}, \textit{RetinaNet}\footnote{\url{https://github.com/facebookresearch/detectron2}\label{detectron2}}
~\cite{lin2017focal}, \textit{YOLO-V5}\footnote{\url{https://github.com/ultralytics/yolov5}\label{yolov5}}
~\cite{glenn_jocher_2022_6222936}, \textit{EfficientDet}\textsuperscript{\ref{nvidia_joc}}
~\cite{tan2020efficientdet} are chosen as the experiment target models. For the segmentation task, \textit{Mask R-CNN}\textsuperscript{\ref{detectron2}}
~\cite{he2017mask} is chosen. \emph{\textbf{RN50}} 
in the brackets represent the ResNet-50 
model served as the backbone of the detection and segmentation models. \emph{\textbf{AP}} represents the box average precision metric for detection task and mask average precision metric for segmentation task.


\subsection{Hyper-Parameters in Experiments}
\label{subsec:hyperparameters}

For the classification, object detection and segmentation networks, we follow the hyper-parameters settings in public repositories marked by the footnotes and detailed list in Table~\ref{Table.hyperparams}. Multiple V100~\cite{nvidiav100} or A100~\cite{nvidiaa100} GPUs are used for data-parallel training in each training, fine-tuning or defense experiment.



\subsection{Comparison Experiments on Benign Dataset}
\label{subsec:comparison_benign_dataset}
\begin{figure}[!htb]
\vskip -0.1in
\centering
    \begin{minipage}[b]{\linewidth}
		\centering
		\includegraphics[width = 0.99\linewidth]{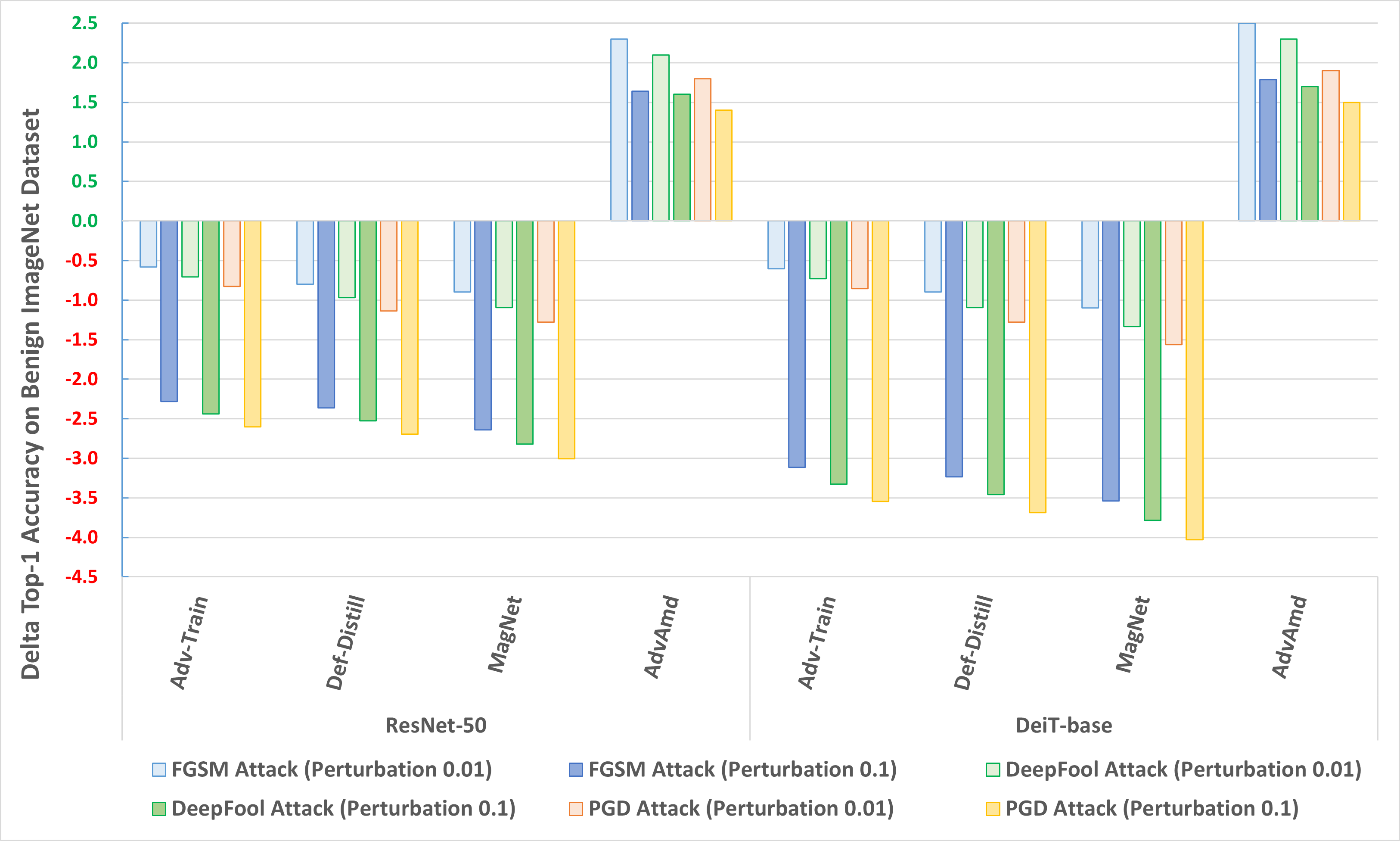}
	\end{minipage}
\vskip -0.1in
	\caption{Comparison on benign \textit{ImageNet} dataset. Only \textit{AdvAmd} solves the accuracy degradation. (The variance is within $\pm$0.04 for \emph{\textbf{Top-1}} with different random seeds.)}
	\label{Figure.2}
\vskip -0.10in
\end{figure}

\begin{figure*}[htb]
\centering
    \begin{minipage}[b]{\linewidth}
		\centering
		\includegraphics[width = 0.99\linewidth]{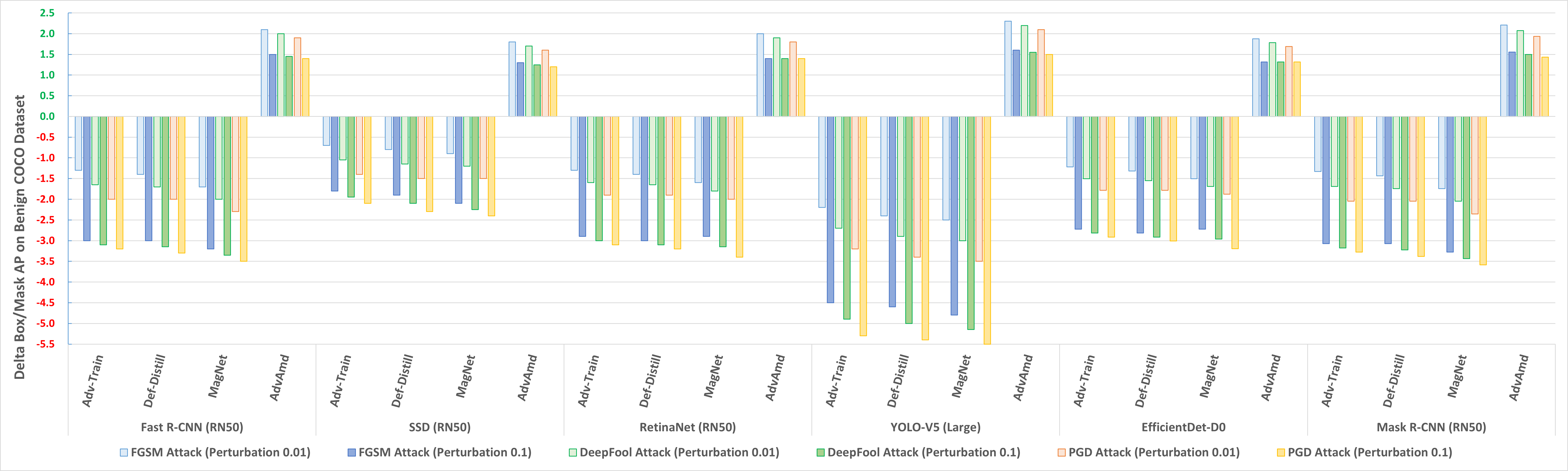}
	\end{minipage}
	\caption{Comparison on benign \textit{COCO} dataset. Only \textit{AdvAmd} solves the accuracy degradation. (The variance is within $\pm$0.07 for box/mask \emph{\textbf{AP}} with different random seeds.)}
	\label{Figure.3}
\end{figure*}

To confirm whether the proposed \textit{AdvAmd} method is effective in solving the accuracy degradation for benign samples, we apply it along with the representative defense methods: adversarial training~\cite{szegedy2013intriguing} (\textit{Adv-Train}), defensive distillation~\cite{papernot2017extending} (\textit{Def-Distill}), and \textit{MagNet}~\cite{meng2017magnet} to 
various models, and test the corresponding accuracy on the benign \textit{ImageNet}~\cite{deng2009imagenet} and \textit{COCO}~\cite{lin2014microsoft} test dataset. To be clearer, only the delta Top-1 accuracy and box/mask average precision metrics are shown in Figure~\ref{Figure.2} and Figure~\ref{Figure.3}. The loss adjustment parameters ($\beta_1, \beta_2, \beta_3$) among the loss items of benign, mediate and adversarial samples all apply value 1.0. We apply a fixed mediate coefficient $\varphi$ value for each adversarial attack, i.e., 0.7 for \textit{FGSM}, 0.6 for \textit{DeepFool} and 0.5 for \textit{PGD}.

Compared to the prior art of the adversarial defense methods with negative delta 
metrics, only the \textit{AdvAmd} method solves the accuracy degradation on the benign dataset. With all three adversarial attacks, \textit{AdvAmd} method has better accuracy boosting performance when the adversarial perturbation is lower. Because the more significant perturbations lead to more severe distribution drift and mismatch between the adversarial and benign samples\footnote{In these and following experiments, we always use the \textit{BN} trained for benign examples when testing the amended model. There are two reasons. Firstly, the amended model does not know whether the input is a benign or adversarial example during testing. So the proper behavior is to treat each test input as a benign example. If the model assumes that test input is an adversarial example and applies the \textit{BN} trained for adversarial examples, even though it may have good robustness, it will be less convincing as the assumption is over-estimated. Secondly, to make a fair comparison with the other prior arts, we can only use the \textit{BN} trained for benign examples because there is no auxiliary \textit{BN} 
in the other methods.}.


\subsection{Comparison Experiments on Adversarial Set}
\label{subsec:comparison_adversarial_dataset}
\begin{figure}[!htb]
\centering
    \begin{minipage}[b]{\linewidth}
		\centering
		\includegraphics[width = 0.99\linewidth]{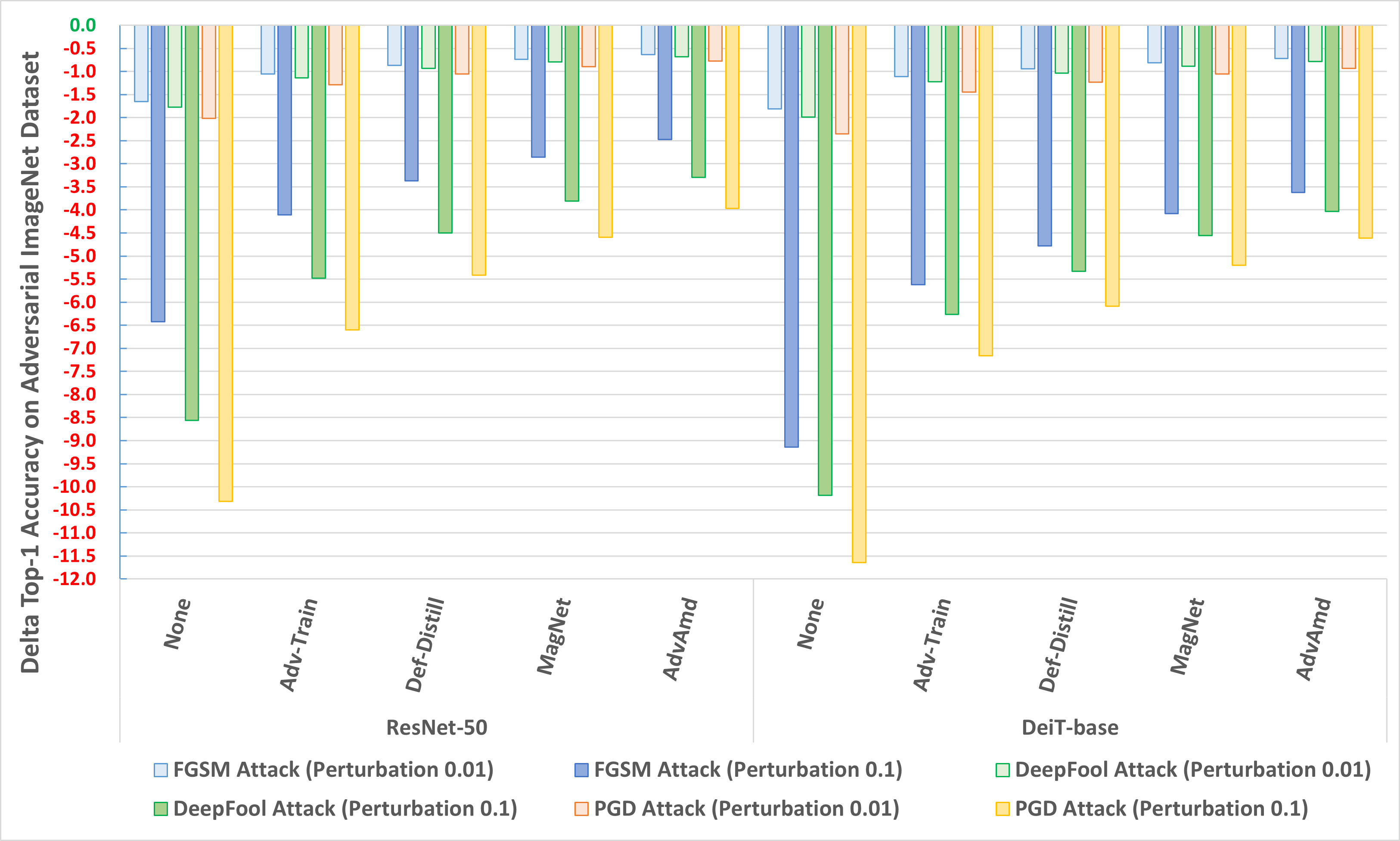}
	\end{minipage}
	\caption{Results on adversarial \textit{ImageNet} dataset. \textit{AdvAmd} remains the adversarial defense capability. (The variance is within $\pm$0.06 for \emph{\textbf{Top-1}} with different random seeds.)}
	\label{Figure.4}
\end{figure}

\begin{figure*}[htb]
\centering
    \begin{minipage}[b]{\linewidth}
		\centering
		\includegraphics[width = 0.99\linewidth]{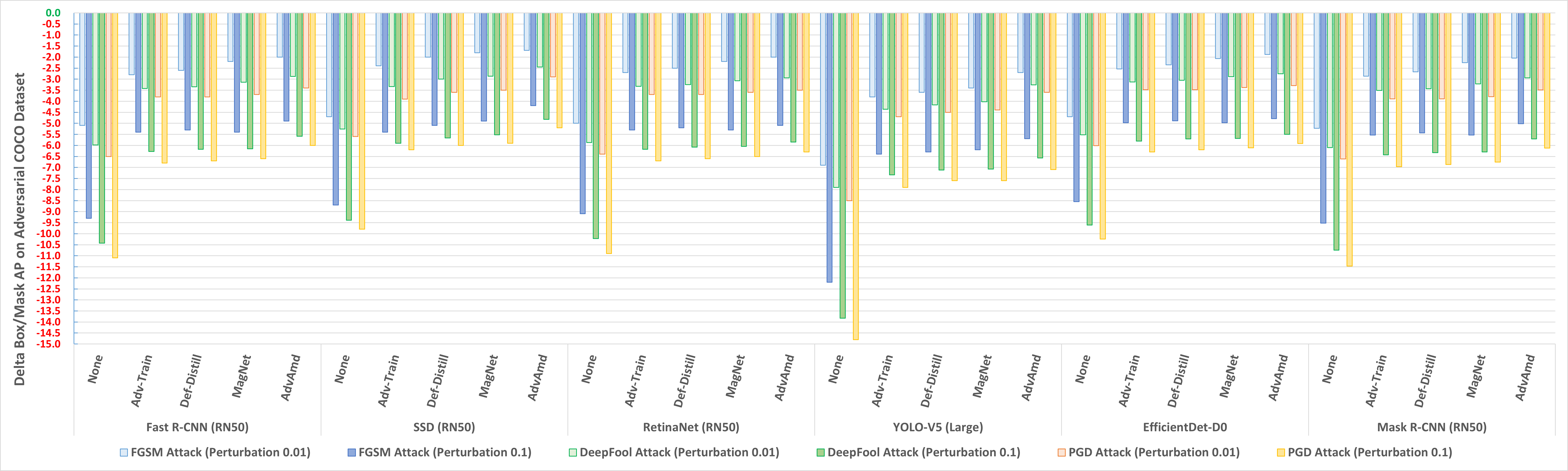}
	\end{minipage}
	\caption{Results on adversarial \textit{COCO} dataset. \textit{AdvAmd} remains the adversarial defense capability. (The variance is within $\pm$0.15 for box/mask \emph{\textbf{AP}} with different random seeds.)}
	\label{Figure.5}
\end{figure*}

To confirm whether 
\textit{AdvAmd} method is still effective as a defense strategy, we repeat the same experiment setting as the previous section
, while testing the corresponding accuracy on the adversarial attacked \textit{ImageNet} and \textit{COCO} dataset. To be clearer, only the delta Top-1 accuracy and box/mask average precision metrics are shown in Figure~\ref{Figure.4} and Figure~\ref{Figure.5}.

Compared to the 
metrics degradation on the adversarial dataset without adversarial defense involved, the \textit{AdvAmd} method 
and the prior art of the adversarial defense methods, can compensate and reduce the accuracy gap. That proves the \textit{AdvAmd} method can still defend the adversarial attacks and improve the adversarial robustness of the 
neural models. By decreasing the distribution mismatch between the benign and adversarial samples and disentangling the mutual learning with an auxiliary \textit{BN} mechanism, the judgment distribution of the amended model is more centralized to the mean of each category. So the vague regions between the adjacent categories will be reduced. That's why 
\textit{AdvAmd} 
can maintain good performance as an adversarial defense strategy.

\subsection{Ablation Experiments and Insights}
\label{subsec:ablation_experiments}
In this experiment, we want to check the contribution of each key component in \textit{AdvAmd} to the final accuracy boosting performance on the benign dataset, as well as the adversarial defense efficacy. Then we can have a deep insight into why \textit{AdvAmd} can harness and transfer the adversarial attack and whether we can further improve it. We include three key components which may have an apparent potential contribution, i.e., the utilization of \textbf{1. mediate samples}, \textbf{2. auxiliary \textit{BN}}, \textbf{3. \textit{AdvAmd} loss}.

\begin{table}[h!]
\centering
\resizebox{0.995\linewidth}{!}{
\begin{tabular}{lccccccccccc}
\toprule
\multirow{3}{*}{\textbf{Network}} & \multirow{3}{*}{\textbf{\begin{tabular}[c]{@{}c@{}}Mediate\\ Samples\end{tabular}}} & \multirow{3}{*}{\textbf{\begin{tabular}[c]{@{}c@{}}Auxiliary\\ Batch Norm\end{tabular}}} & \multirow{3}{*}{\textbf{\begin{tabular}[c]{@{}c@{}}\textit{AdvAmd}\\ Loss\end{tabular}}} & \multicolumn{4}{c}{\textbf{$\Delta$ Mask/Box AP on Adversarial Dataset}} & \multicolumn{4}{c}{\textbf{$\Delta$ Mask/Box AP on Benign Dataset}} \\
\cmidrule(lr){5-8}\cmidrule(lr){9-12}
 & & & & \multicolumn{2}{c}{\textbf{FGSM Attack}} & \multicolumn{2}{c}{\textbf{PGD Attack}} & \multicolumn{2}{c}{\textbf{FGSM Attack}} & \multicolumn{2}{c}{\textbf{PGD Attack}}\\
\cmidrule(lr){5-6}\cmidrule(lr){7-8}\cmidrule(lr){9-10}\cmidrule(lr){11-12}
 & & & & \multicolumn{1}{c}{\textbf{$\epsilon$=0.01}} & \multicolumn{1}{c}{\textbf{$\epsilon$=0.1}} & \multicolumn{1}{c}{\textbf{$\epsilon$=0.01}} & \multicolumn{1}{c}{\textbf{$\epsilon$=0.1}} & \multicolumn{1}{c}{\textbf{$\epsilon$=0.01}} & \multicolumn{1}{c}{\textbf{$\epsilon$=0.1}} & \multicolumn{1}{c}{\textbf{$\epsilon$=0.01}} & \multicolumn{1}{c}{\textbf{$\epsilon$=0.1}} \\
\midrule
\multirow{7}{*}{DeiT-base} 
 & \Checkmark   & \XSolidBrush & \XSolidBrush   & -1.6  & -7.2   & -2.1  & -9.2   & -1.3  & -2.6  & -2.8  & -3.6 \\
 & \XSolidBrush & \Checkmark   & \XSolidBrush   & -1.4  & -7.1   & -1.9  & -8.7   &  0.4  &  0.1  &  0.3  &  0.1 \\
 & \XSolidBrush & \XSolidBrush & \Checkmark     & -1.4  & -7.0   & -1.8  & -8.5   &  0.5  &  0.2  &  0.4  &  0.1 \\
 & \Checkmark   & \Checkmark   & \XSolidBrush   & -1.2  & -4.2   & -1.5  & -5.3   &  1.4  &  0.8  &  1.1  &  0.6 \\
 & \Checkmark   & \XSolidBrush & \Checkmark     & -1.2  & -4.1   & -1.4  & -5.2   &  1.5  &  0.9  &  1.2  &  0.7 \\
 & \XSolidBrush & \Checkmark   & \Checkmark     & -1.1  & -3.9   & -1.2  & -4.8   &  2.3  &  1.6  &  1.8  &  1.2 \\
 & \Checkmark   & \Checkmark   & \Checkmark     & -0.7  & -3.6   & -0.9  & -4.6   &  2.5  &  1.8  &  1.9  &  1.5 \\
\midrule
\multirow{7}{*}{\begin{tabular}[c]{@{}l@{}}Mask R-CNN \\ (RN50)\end{tabular}} 
 & \Checkmark   & \XSolidBrush & \XSolidBrush   & -3.0  & -6.1   & -4.9  & -7.5   & -2.4  & -4.8  & -3.3  & -5.4 \\
 & \XSolidBrush & \Checkmark   & \XSolidBrush   & -2.6  & -5.5   & -4.3  & -6.7   & -0.1  & -1.0  & -0.3  & -1.1 \\
 & \XSolidBrush & \XSolidBrush & \Checkmark     & -2.6  & -5.6   & -4.4  & -6.9   & -0.1  & -1.1  & -0.3  & -1.2 \\
 & \Checkmark   & \Checkmark   & \XSolidBrush   & -2.3  & -5.4   & -3.8  & -6.3   &  1.2  &  0.2  &  1.0  &  0.2 \\
 & \Checkmark   & \XSolidBrush & \Checkmark     & -2.4  & -5.4   & -4.0  & -6.5   &  1.1  &  0.1  &  0.9  &  0.1 \\
 & \XSolidBrush & \Checkmark   & \Checkmark     & -2.2  & -5.3   & -3.8  & -6.4   &  1.9  &  1.3  &  1.7  &  1.2 \\
 & \Checkmark   & \Checkmark   & \Checkmark     & -2.0  & -5.0   & -3.5  & -6.1   &  2.2  &  1.6  &  1.9  &  1.4 \\
\midrule
\multirow{7}{*}{\begin{tabular}[c]{@{}l@{}}YOLO-V5\\ (Large)\end{tabular}} 
 & \Checkmark   & \XSolidBrush & \XSolidBrush   & -3.6  & -6.5   & -4.6  & -8.0   & -2.6  & -5.0  & -3.7  & -5.8 \\
 & \XSolidBrush & \Checkmark   & \XSolidBrush   & -3.3  & -6.1   & -4.2  & -7.7   &  0.1  & -1.0  & -0.3  & -1.2 \\
 & \XSolidBrush & \XSolidBrush & \Checkmark     & -3.4  & -6.3   & -4.4  & -7.9   &  0.1  & -1.2  & -0.4  & -1.4 \\
 & \Checkmark   & \Checkmark   & \XSolidBrush   & -3.0  & -5.9   & -3.9  & -7.4   &  1.2  &  0.2  &  1.0  &  0.1 \\
 & \Checkmark   & \XSolidBrush & \Checkmark     & -3.1  & -6.0   & -4.1  & -7.6   &  1.0  &  0.1  &  0.7  &  0.0 \\
 & \XSolidBrush & \Checkmark   & \Checkmark     & -3.0  & -5.9   & -3.9  & -7.3   &  1.9  &  1.3  &  1.8  &  1.2 \\
 & \Checkmark   & \Checkmark   & \Checkmark     & -2.7  & -5.7   & -3.6  & -7.1   &  2.3  &  1.6  &  2.1  &  1.5 \\
\bottomrule
\end{tabular}
}
\caption{Ablation experiment to check the contribution of key components in \textit{AdvAmd} method on detection task.}
\label{Table.2}
\end{table}

We make the combination to enable or disable among three key components
. The ablation results are shown in Table~\ref{Table.2}. Suppose we only apply the \textit{AdvAmd} method with the lacking of one key component. In that case, we can find the absence of the mediate samples has about $\textbf{0.1\%}\sim\textbf{0.3\%}$ decrease of \emph{\textbf{Top-1}} accuracy for \textit{DeiT-base}, $\textbf{0.2\%}\sim\textbf{0.3\%}$ decrease of the mask \emph{\textbf{AP}} for \textit{Mask R-CNN} and $\textbf{0.3\%}\sim\textbf{0.4\%}$ decrease of the box \emph{\textbf{AP}} for \textit{YOLO-V5} on the benign dataset, while the absence of the utilization of the auxiliary \textit{BN} or \textit{AdvAmd} loss has around $\textbf{0.7\%}\sim\textbf{1.1\%}$ \emph{\textbf{Top-1}} accuracy decrease for \textit{DeiT-base}, $\textbf{1.1\%}\sim\textbf{1.5\%}$ mask accuracy decrease for \textit{Mask R-CNN} and $\textbf{1.1\%}\sim\textbf{1.5\%}$ box accuracy decrease for \textit{YOLO-V5} on the benign dataset. It seems that the contribution of the mediate samples is less than the other two components. We further verify the ablation results with the \textit{AdvAmd} method, which only applies one of the key components. If only enabling the mediate samples for the \textit{AdvAmd} method, the accuracy degradation on the benign dataset is still severe. While applying one of the auxiliary \textit{BN} or \textit{AdvAmd} loss, the accuracy degradation will be healed. The delta box \emph{\textbf{AP}} on the benign dataset between the amended and vanilla target models is close to zero for the adversarial attack with minor perturbations. So we can confirm that \textbf{\textit{the auxiliary \textit{BN} and \textit{AdvAmd} loss are two powerful improvements to heal the accuracy degradation on the benign dataset.}} However, we should notice that if the utilization of the mediate samples is enabled with one of the auxiliary \textit{BN} or \textit{AdvAmd} loss, there is an obvious accuracy boost. That proves \textbf{\textit{the separate utilization of the mediate samples can help in a relatively small margin}}. While at the same time, \textbf{\textit{the combination utilization of the mediate samples with the other two amendment components can also help in a relatively large margin}}.

On the other hand, 
the lacking of one or two key components only leads to a limited negative influence on the 
defense performance on the adversarial dataset. These three key components are 
designed to solve the accuracy degradation of the benign dataset caused by distribution mismatch and 
mutual learning
, so the limited influence on the defense efficacy is not surprising. Moreover, the prior art defense strategy's principle is teaching the target model that the adversarial samples in the vague judgment range can still be regarded as the normal samples. While the key components introduced in \textit{AdvAmd} teach the amended model to judge the benign, mediate, and adversarial samples in finer granularity. From the 
results on the adversarial dataset, the finer-granularity division for 
these samples can only provide some but limited help for the adversarial defense efficacy. It reveals an interesting conclusion that \textbf{\textit{the finer granularity uses a sled-hammer on a gnat for 
adversarial defense task.}}

\section{Limitations and Future Work}
In the 
\textit{AdvAmd} method, we mainly consider the adversarial attack pattern to mislead the models in categories detection and classification. Although this is the prior art attack strategy, whether the \textit{AdvAmd} method can also help to harness the attack aiming at confusing the detected and segmented positions has not been demonstrated. We leave it for further study in the future work.


\section{Conclusions}
\label{sec:conclusion}
We notice the prior art adversarial defense methods lead to accuracy degradation of the classification, object detection and segmentation models on the benign dataset. Based on the qualitative explanation and theoretical proof, we find the distribution mismatch between the benign and adversarial samples and the mutual learning mechanism with same learning ratio applied in prior art adversarial defense strategies is the root cause. Then we propose the \textit{AdvAmd} method to harness the adversarial attack for healing the accuracy degradation. Three key components: \textbf{mediate samples} (to reduce the influence of distribution mismatch with a fine-grained amendment), \textbf{auxiliary batch norm} (to solve the mutual learning mechanism and the smoother judgment surface), and \textbf{\textit{AdvAmd} loss} (to adjust the learning ratios according to different attack vulnerabilities) make the main contribution to amending the adversarial attack in the right manner.


\appendix

\section*{Ethical Statement}

We should emphasize the aim of \textit{AdvAmd} is transferring and healing the adversarial attacks' influence on various classification and object detection tasks. However, we encourage the community to understand and mitigate the risks arising from the \textit{AdvAmd} method. As the principle and the implementation of \textit{AdvAmd} will be public, people may study the novel adversarial attack to deactivate the \textit{AdvAmd} intentionally. We should notice the risk that \textit{AdvAmd} is misused to help evolve more powerful attacks that can be used to misrepresent objective truth. 

\section*{Acknowledgements}

This work is supported by National Natural Science Foundation of China (No. 62071127, U1909207 and 62101137), Shanghai Municipal Science and Technology Major Project (No.2021SHZDZX0103), Shanghai Natural Science Foundation (No. 23ZR1402900), Zhejiang Lab Project (No. 2021KH0AB05).

\bibliographystyle{named}
\bibliography{ijcai23}

\end{document}